\documentclass[11pt,a4paper]{article}

\usepackage[T2A,T1]{fontenc}
\usepackage[utf8]{inputenc}
\usepackage[russian,english]{babel}
\usepackage{newtxtext}
\usepackage{newtxmath}
\usepackage{substitutefont}
\substitutefont{T2A}{\rmdefault}{Tempora-TLF}

\usepackage[hyperref]{eacl2021}
\usepackage{times}
\usepackage{latexsym}

\usepackage{microtype}
\usepackage{booktabs}
\usepackage{graphicx}
\usepackage{multirow}
\usepackage{todonotes}
\usepackage{xcolor}
\usepackage{titlesec}

\colorlet{red}{black}

\titlespacing{\paragraph}{%
  1pt}{
  0.2\baselineskip}{
  1em}

\aclfinalcopy 

\newcommand{\mn}[1]{\textsc{#1}}

\title{Enhancing Sequence-to-Sequence Neural Lemmatization with External Resources}

\author{Kirill Milintsevich \\
  Institute of Computer Science \\
  University of Tartu \\
  Tartu, Estonia \\
  \texttt{kirill.milintsevich@ut.ee} \\\And
  Kairit Sirts \\
  Institute of Computer Science \\
 University of Tartu \\
  Tartu, Estonia \\
  \texttt{kairit.sirts@ut.ee} \\}

\date{}

\begin{document}
\maketitle
\begin{abstract}
We propose a novel hybrid approach to lemmatization\footnote{\url{https://github.com/501Good/lexicon-enhanced-lemmatization}} that enhances the seq2seq neural model with additional lemmas extracted from an external lexicon or a rule-based system.
During training, the enhanced lemmatizer learns both to generate lemmas via a sequential decoder and copy the lemma characters from the external candidates supplied during run-time.
Our lemmatizer enhanced with candidates extracted from the Apertium morphological analyzer achieves statistically significant improvements compared to baseline models not utilizing additional lemma information, achieves an average accuracy of 97.25\% on a set of 23 UD languages, which is 0.55\% higher than obtained with the Stanford Stanza model on the same set of languages.
We also compare with other methods of integrating external data into lemmatization and show that our enhanced system performs considerably better than a simple lexicon extension method based on the Stanza system, and it achieves complementary improvements w.r.t. the data augmentation method.
\end{abstract}

\section{Introduction}

State-of-the-art lemmatization systems are based on attentional sequence-to-sequence neural architectures operating on characters that transform the surface word form into its lemma \citep{kanerva2018turku,qi2018universal}.
Like any other supervised learning model, these systems are dependent on the amount and quality of the existing training data.
Attempts to develop even more accurate lemmatization systems can focus on improving the model's architecture or obtaining additional data. While annotating additional data is an ongoing process for many smaller languages in the Universal Dependencies (UD) collection, there are also other data sources available that can be useful for improving lemmatization systems. In particular, we refer to existing rule-based morphological analyzers, lexicons, and other such resources.

Three potential sources for extracting additional lemma candidates are Apertium, Unimorph, and UD Lexicons initiatives. Apertium\footnote{\url{https://www.apertium.org}} is an open-source rule-based machine translation platform \citep{forcada2011apertium}.
It also includes rule-based morphological analyzers based on finite-state transducers that cover 80 languages. Unimorph\footnote{\url{http://unimorph.org/}} is a project aimed at collecting annotated morphological inflection data, including lemmas, from Wiktionary \citep{KIROV16.1077}, a free open dictionary for many languages. Currently, the Unimorph project covers 110 languages. UD Lexicons\footnote{\url{http://atoll.inria.fr/~sagot/}} is a collection of 53 morphological lexicons in CoNLL-UL format covering 38 languages. UD Lexicons mostly use Apertium and Giellatekno systems to generate the annotations~\citep{sagot2018multilingual}.

Several previous works have proposed methods to improve lemmatization systems by augmenting the training data with additional instances \citep{bergmanis2019training,kanerva2020universal}. In this paper, we propose another approach that both modifies the model architecture and leverages additional data. Unlike previous work where the model gains from extracting extra knowledge from the additional data provided for training, our primary goal is to teach the model to use external resources, even those that may only be available later during test time. 
In particular, \textcolor{red}{the proposed system is a dual-encoder model, which} receives two inputs for each word: 1) the word form itself to be lemmatized and 2) (optionally) the lemma candidates for that word form extracted from a lexicon or generated by a rule-based system. Both inputs are encoded with two different encoders and passed to the decoder. The decoder then learns via two separate attentional mechanisms to generate the lemma via \textcolor{red}{the combination of the regular transduction and by} copying characters from the external candidates. This way, the model is trained to use two sources of information--the regular training set and the options proposed by an external resource.

\textcolor{red}{The experiments with several models enhanced with external data on 23 UD languages show that the best model using} additional lemma candidates generated by the Apertium system achieves significantly higher results than the baseline models trained on the UD training set only. Also, we compare our method with other methods using external data. \textcolor{red}{The} enhanced system performs considerably better than a simple lexicon extension method based on the Stanza system, and it achieves complementary improvements w.r.t. the data augmentation method of \citet{kanerva2020universal}.

\section{Related Works}
\label{sec:related}

Nowadays, state-of-the-art lemmatization systems are typically based on a neural sequence-to-sequence architecture, as demonstrated by the variety of systems presented at the
CoNLL 2018 \citep{zeman2018conll} and SIGMORPHON 2019 \citep{mccarthy2019sigmorphon} shared tasks.
Several systems, including the TurkuNLP pipeline, the winner of the lemmatization track at CoNLL 2018 Shared task, use an attention-based translation model \citep{kanerva2018turku,qi2018universal}. The input to the system is the character sequence of a surface form (SF), which is ``translated" into the lemma by an attention-based decoder. The input sequence can also be extended with POS tags \citep{qi2018universal} and morphological features \citep{kanerva2018turku}.

Another approach was used by the UDPipe Future system, the second-best model at the CoNLL 2018 Shared Task. \citet{straka2018udpipe} proposed to produce a lemma by constructing a set of rules that transform the SF into a lemma. These rules can include copying, moving, or deleting a character in the SF, as well as additional rules for changing or preserving the casing. Thus, the lemmatization task is rendered into a multi-class classification task of choosing the correct transformation rule among the set of all possible rules generated from the training set. A year later, \citet{straka2019udpipe} improved the result for the lemmatization by adding BERT contextual embeddings \citep{devlin2019bert} to the input, which made them the best lemmatization system at the SIGMORPHON 2019 Shared Task. 

Several previous works have proposed to leverage additional data to improve lemmatization. In the simplest form, training data itself can be used to create a lexicon that maps word forms to its lemma. This strategy has been adopted by the Stanford neural lemmatization system \citep{qi2018universal}, which creates such lexicons from the training sets and resorts to lemma generation only when the lexicon lookup fails. One can easily imagine extending such a lexicon with external resources. \citet{rosa2018cuni} adopted another simple way of using Unimorph lexicons to post-fix the morphological features and lemmas predicted by the UDPipe system \citep{straka-strakova-2017-tokenizing}. The post-fix is performed by simply looking up the SF from the Unimorph lexicon and, if the match is found, replacing the model prediction with the tags and lemmas found in the lexicon.

Another line of work has used additional data to augment the training data set. \citet{bergmanis2019training} augmented their training set by first listing all non-ambiguous word-lemma pairs from Unimorph lexicons and then extracted sentences from Wikipedia that contained these words. They then trained the context-sensitive Lematus model \citep{bergmanis2018context} on this extended partially lemmatized data set.
\citet{kanerva2018turku} used Apertium's morphological analyzer module to extend the training set for languages with tiny UD datasets. Apertium was used to generate all possible morphological analyses to 5000 sentences selected from the Wikipedia of the respective language. For each sentence, the most likely analysis sequence was then obtained via a disambiguating language model. \textcolor{red}{The words that were assigned an Apertium-generated lemma during this process were added to the lemmatizer training set.}
\textcolor{red}{In the subsequent work, \citet{kanerva2020universal} extended the training data even more. They used Apertium to analyze all words found in the CoNLL 2017 web crawl dataset \citep{ginter2017conll} or in the Wikipedia of the respective language. All new words with unambiguous lemma and morphological analysis were added to the augmented training set.}

\section{Method}
\label{sec:method}

\begin{figure}[t]
\includegraphics[width=0.48\textwidth]{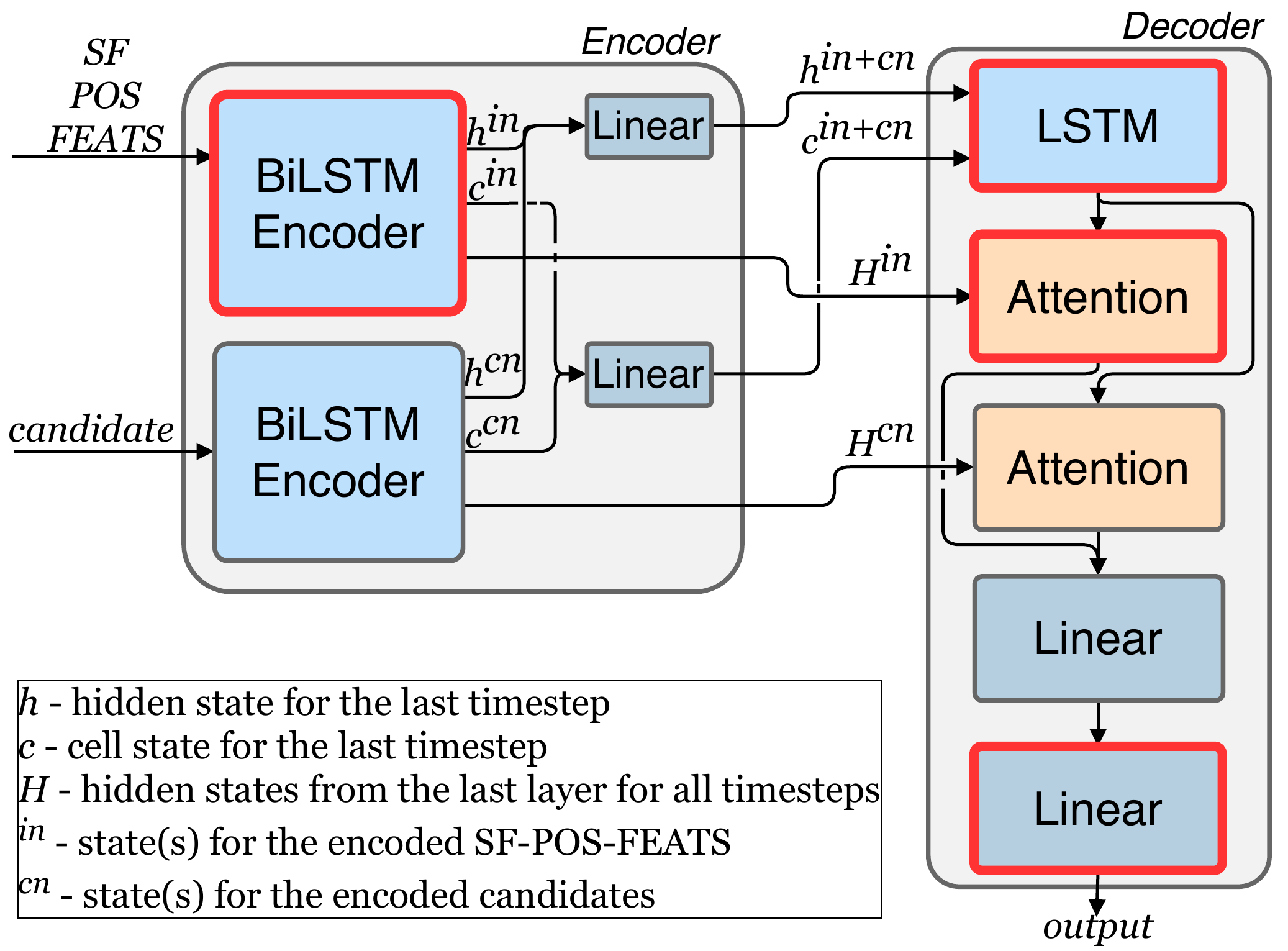}
\caption{The architecture of the dual-encoder enhanced lemmatizer. Layers that comprise the original Stanza lemmatizer are marked with a bold red border.}
\label{fig:model_general}
\end{figure}

\textcolor{red}{The core of the proposed model is the Stanford lemmatizer \citep{qi2018universal,qi2020stanza} which is a sequence-to-sequence model with attention.} It takes character-level word representation and the POS tag as input and processes them with a bidirectional LSTM encoder. Then, it passes the encoder outputs to an LSTM decoder, which applies a soft dot attention layer after every LSTM cell. Finally, the output is constructed via greedy decoding.

We make several changes to the model architecture as shown in Figure~\ref{fig:model_general}. The components comprising the original Stanford lemmatizer are marked on the figure with the bold red border. First, we add another encoder that encodes the lemma candidates provided by the external system. The output representations of both encoders are combined with a linear layer and fed to the decoder. Secondly, we add another attention layer to the decoder that attends to the outputs of the second encoder. 
The outputs are finally combined with a linear layer.
Finally, in addition to the POS tag, we also add morphological features to the first encoder's input. 

Additionally, we implement the encoder dropout to simulate the situation when the \textcolor{red}{external} candidates are absent. The value of the encoder dropout that varies in the range of $\{0.0, 1.0\}$ defines the probability of discarding \textcolor{red}{all candidates from a batch during training.} Thus, the model will train only the main encoder based on this batch. This helps to train the model to perform more robustly in both situations when the candidates in the second encoder are present or absent.

\begin{table*}[t]
\centering
\setlength\tabcolsep{5 pt}
\begin{tabular}{l r | c c c c c | c c r r} 
 \toprule
 \multirow{2}{*}{\bf Treebank} & \multirow{2}{1cm}{\bf Size} & \multicolumn{5}{c}{\bf All words} & \multicolumn{4}{c}{\bf Out-of-vocabulary}\\
  & & \bf \mn{Def} & \bf \mn{Lex} & \bf \mn{Uni} & \bf \mn{Apt} & \bf \mn{Stanza} & \bf \mn{Def} & \bf \mn{Apt} & \bf Diff & \bf OOV\%\\[0.5ex] 
\midrule
\midrule
 cs\_pdt & 1,503K & 98.51 & 98.66 & \bf 98.67 & 98.55 & 98.58 & 90.51 & 90.95 & 0.44 & 7.53\\ 
 ru\_syntagrus & 1,107K & 97.82 & 97.92 & 98.00 & \bf 98.11 & 97.91 & 89.48 & 91.65 & 2.17 & 10.56\\
 es\_ancora & 547K & 99.31 & 99.28 & 99.31 & \bf 99.35 & 99.21 & 95.06 & 95.40 & 0.34 & 5.90\\
 ca\_ancora & 530K & 98.85 & 98.83 & 98.83 & \bf 98.89 & 98.49 & 95.82 & 96.79 & 0.97 & 5.43\\
 fr\_gsd & 389K & 98.05 & 98.07 & 98.10 & 98.13 & \bf 98.15 & 89.50 & 90.83 & 1.33 & 6.19\\
 hi\_hdtb & 351K & 98.77 & 98.71 & 98.71 & \bf 98.80 & 96.66 & 93.49 & 94.62 & 1.13 & 4.67\\
 de\_gsd & 287K & 96.87 & 96.91 & \bf 97.04 & 96.80 & 96.78 & 85.53 & 85.22 & -0.31 & 13.04\\
 it\_isdt & 278K & 98.19 & 98.39 & 98.31 & \bf 98.48 & 98.32 & 90.28 & 92.22 & 1.94 & 5.86\\
 en\_ewt & 254K & 98.21 & 98.19 & 98.22 & \bf 98.26 & 98.18 & 90.10 & 90.49 & 0.39 & 10.05\\
 ro\_rrt & 218K & 98.33 & 98.28 & 98.32 & \bf 98.53 & 98.16 & 91.46 & 93.22 & 1.76 & 11.60\\
 pt\_bosque & 210K & 98.24 & 98.20 & 98.23 & \bf 98.32 & 98.12 & 93.15 & 94.27 & 1.12 & 8.85\\
 nl\_alpino & 208K & \bf 97.08 & 96.61 & 96.89 & 96.74 & 96.99 & 86.34 & 84.88 & -1.46 & 15.81\\
 bg\_btb & 156K & 97.97 & \bf 98.20 & 98.17 & 98.07 & 97.36 & 91.07 & 91.02 & 0.05 & 13.97 \\
 ur\_udtb & 138K & 97.16 & \bf 97.29 & 97.28 & 97.28 & 95.62 & 91.83 & 91.93 & 0.01 & 6.79\\
 gl\_ctg & 126K & 98.48 & 98.48 & 98.51 & \bf 98.93 & 98.59 & 89.73 & 93.55 & 3.82 & 10.94\\
 uk\_iu & 122K & 97.03 & 97.07 & 97.06 & \bf 97.12 & 96.70 & 91.15 & 91.32 & 0.17 & 33.62\\
 eu\_bdt & 121K & 96.48 & 96.62 & 96.63 & \bf 96.68 & 96.52 & 86.18 & 86.81 & 0.63 & 21.68\\
 da\_ddt & 100K & 97.87 & 97.7 & 97.81 & \bf 98.03 & 97.36 & 89.86 & 90.31 & 0.45 & 18.13\\
 sv\_talbanken & 96K & 97.36 & 97.59 & 97.64 & \bf 98.27 & 97.53 & 87.66 & 92.33 & 4.67 & 17.52\\
 el\_gdt & 61K & 96.84 & 97.06 & 97.25 & \bf 97.38 & 96.66 & 84.18 & 86.42 & 2.24 & 19.59\\
 tr\_imst & 56K & 97.03 & 97.23 & 97.13 & \bf 97.39 & 96.73 & 92.27 & 93.18 & 0.91 & 36.25\\
 hy\_armtdp & 52K & 95.55 & 95.84 & 94.87 & \bf 96.01 & 95.55 & 86.11 & 87.34 & 1.23 & 38.54\\
 be\_hse & 13K & 81.91 & 81.86 & 82.36 & \bf 82.63 & 79.98 & 68.78 & 70.30 & 1.52 & 93.28\\
\midrule
 Average & & 97.04 & 97.09 & 97.10 & \bf 97.25 & 96.70 & 89.11 & 90.22 & 1.11 \\[1ex] 
\bottomrule
\end{tabular}
\caption{Lemmatization accuracy of the models enhanced with training the set lexicon (\mn{Lex}), Unimorph lexicon (\mn{Uni}), and Apertium systems (\mn{Apt}) as well as the \mn{Default} (\mn{Def})and \mn{Stanza} baselines on 23 UD languages.}
\label{tab:main_results}
\vspace{-4mm}
\end{table*}

\section{Experiments}
\label{sec:experiments}

\paragraph{Data}

\textcolor{red}{The models are trained and tested on the Universal Dependencies (UD) v2.5 corpora \citep{11234/1-3105_short}. As additional external data, the lexicons from the Unimorph project \citep{KIROV16.1077}, UD Lexicons \citep{sagot2018multilingual}, and lemmas generated with the Apertium morphological analyzer module \citep{forcada2011apertium} are used.}
 \textcolor{red}{We also experiment with the lexicon constructed from the training set to simulate the situation when no additional data is available---this scenario assesses the effect of the second encoder without external data.
The experiments are conducted on 23 languages from the UD collection. The basis of this selection was that all these languages are supported by both Unimorph, UD Lexicons, and Apertium.}

To extract lemmas from the Unimorph lexicon, the input \textcolor{red}{surface form} (SF) is queried from the lexicon to retrieve the corresponding lemma. Some morphological forms in the Unimorph lexicons consist of several space-separated tokens; these were discarded.
UD Lexicons are presented in the CoNLL-UL format, which is an extension of the CoNLL-U format. This makes the extraction process trivial since the lexicons are already pretokenized.
For Apertium, all generated lemmas were stripped from special annotation symbols, and duplicate lemmas were removed. 
Finally, the simple \textcolor{red}{training set based} lexicon solution, similar to \citet{qi2018universal}, consists of two lookup dictionaries. The first lexicon maps SF-POS pairs to their lemmas, the second lexicon maps just SF's to their possible lemmas found in the training set.
The \textcolor{red}{lemma candidates} for a SF are selected by first querying the input SF and POS tag from the SF-POS dictionary and, in case of failure, falling back to the SF dictionary.

\paragraph{Baselines}
As the first baseline, we compare our results with \mn{Stanza}, the lemmatization module from the Stanza pipeline \cite{qi2020stanza}, which is a repackaging of the Stanford lemmatization system from the CoNLL 2018 Shared Task \citep{qi2018universal}. We used the lemmatization models trained on the UDv2.5 available on the Stanza web page.
\textcolor{red}{As the \mn{Default} baseline, we use our enhanced model, with the second encoder always being empty.}

\paragraph{Experimental Setup}

We train four \textcolor{red}{enhanced dual-encoder} models that differ in the input to the second encoder. For all models, the input to the first encoder is the concatenation of SF characters, POS tag, and morphological features. During the training phase, gold POS tags and morphological features are supplied, while during inference, POS tags predicted with the Stanza tagger are used.
The input to the second encoder is the following: for the second baseline (\mn{Default}), it is always empty; for the \mn{Lexicon}, \mn{Unimorph}, and \mn{Apertium} \textcolor{red}{enhanced} models, it contains the lemma candidate(s) from the training set based lexicon, Unimorph lexicons, and Apertium analyses respectively. If several possible candidates are returned for a SF, then these are concatenated. The encoder dropout for the \mn{Lexicon} model is set to $0.8$ to simulate the situation during testing for \textcolor{red}{out-of-vocabulary (OOV)} words where the second encoder will be empty.
\textcolor{red}{All models were trained in the HPC at the University of Tartu \cite{HPC} for a maximum of $60$ epochs with stopping early if there was no improvement in the development accuracy in $10$ epochs.}
 
\section{Results}

Table~\ref{tab:main_results} shows the results for all three enhanced systems and two baselines. 
The \mn{Apertium} model outperforms other models for most languages, although the absolute differences are quite small. The \mn{Lexicon} model and the \mn{Default} baseline are on the same level on average, suggesting that supplying the model with lemmas extracted from the training set via the second encoder does not help to leverage the training data better. However, all \textcolor{red}{enhanced} models, including the \mn{Default} model, perform better than the \mn{Stanza} baseline, suggesting that omitting the lexicon heuristics and supplying the input tokens with both POS and morphological features might improve performance.

One-way ANOVA was performed to detect statistical difference between the systems.\footnote{The results for be\_hse were extreme outliers and were not included in the comparison. The \mn{Unimorph}\textcolor{red}{-enhanced} model was excluded from this test as its results did not conform to the normality requirement.}  A significant difference between the scores at the $p < 0.05$ level ($p = 0.038$) was found. Post hoc comparisons using one-sided paired t-tests showed that the mean accuracy of the \mn{Apertium}\textcolor{red}{-enhanced} model is significantly greater \textcolor{red}{compared to the}  the \mn{Default} ($p_{adj} = 0.0005$), \mn{Lexicon} ($p_{adj} < 0.0001$), \mn{Unimorph} ($p_{adj} = 0.0001$) and \mn{Stanza} ($p_{adj} < 0.0001$) systems with the $p$-value adjusted for multiple comparisons using the Bonferroni correction.

As the baseline model performances are already very high and the external information is expected to improve the lemmatization \textcolor{red}{most for the} new words \textcolor{red}{unseen during training}, we computed the accuracy of the out-of-vocabulary words (OOV) for the best performing \mn{Apertium} model and the \mn{Default} baseline. In this context, OOV words are those words in the test set that were not seen by the model during training. 
The results are shown in the right-most section of the Table~\ref{tab:main_results}.
\textcolor{red}{The improvements on the OOV words are variable,}
 depending on the language, although on average, the improvement of the \mn{Apertium} model over the \mn{Default} baseline is more than 1\%. We hypothesize that the direction and the magnitude of these effects are dependent on the coverage and the quality of the Apertium \textcolor{red}{morphological} analyzer.

\section{Analysis of the Results}

In this section, we analyze more thoroughly the potential of the proposed method. First, we compare our enhanced system with alternative methods for deploying external data, particularly with the data augmentation method proposed by \citet{kanerva2020universal} and a lexicon extension method implemented based on the Stanza system \cite{qi2020stanza}. Secondly, we present more analyses to provide evidence towards the conclusion that the improvements presented \textcolor{red}{for the enhanced model} in the previous section can be attributed to our system's ability to make use of external resources supplied to the model via the second encoder.

\begin{table*}[t]
\centering
\begin{tabular}{l | c c | c c ||c c c c}
\toprule

\bf Treebank&\bf \mn{Def}&\bf \mn{Apt}&\bf \mn{Def}+8K&\bf \mn{Apt}+8K&\bf \mn{Apt}$_{0.8}$&\bf \mn{Apt}+E&\bf \mn{Apt}+\mn{Uni}&\bf \mn{Apt}+UD\\
& \multicolumn{2}{c|}{Our models} & \multicolumn{2}{c||}{Augmented models} & \multicolumn{4}{c}{The second encoder input varies} \\
\midrule
\midrule
cs\_pdt         & 98.51 & \underline{98.55} & 98.49 & \bf 98.57 & 98.49 & 98.39 & 98.51 & 98.50 \\
ru\_syntagrus   & 97.82 & \bf \underline{98.11} & 97.86 & 98.06 & 97.98 & 97.83 & 97.98 & 97.97\\
es\_ancora      & 99.31 & 99.35 & \underline{99.53} & \bf 99.60 & 99.33 & 99.29 & 99.33 & 99.33\\
ca\_ancora      & 98.85 & \bf \underline{98.89} & 98.86 & \bf 98.89 & 98.85 & 98.80 & 98.85 & 98.85\\
fr\_gsd         & 98.05 & 98.13 & \underline{98.98} & \bf 99.05 & 97.98 & 97.79 & 97.97 & 97.98\\
hi\_hdtb        & 98.77 & 98.80 & \bf \underline{98.83} & 98.78 & 98.84 & 98.66 & 98.83 & 98.84\\
de\_gsd         & \bf 96.87  & \underline{96.80} &96.79&96.67&96.83&96.49&96.83&96.84\\
it\_isdt        & 98.19 & 98.48&\underline{98.98} & \bf 98.99&98.36&98.3&98.36&98.37\\
en\_ewt         & 98.21 & \bf \underline{98.26} &97.24&98.12&98.21&98.17&98.22&98.20\\
ro\_rrt         & 98.33 & \bf \underline{98.53} & 97.56 & 98.48 & 98.44 & 98.29 & 98.46 & 98.41 \\
pt\_bosque      & 98.24 & \bf \underline{98.32} & 98.13 & 98.29 & 98.30 & 98.32 & 98.30 & 98.31 \\
nl\_alpino      & 97.08  & 96.74&\underline{96.80} & \bf 96.82&96.89&96.86&96.81&96.85\\
bg\_btb         & 97.97 & 98.07& \bf \underline{98.84} &98.82&98.02&98.06&98.02&98.02\\
ur\_udtb        & 97.16 & \underline{97.28} &96.90& \bf 97.31&97.13&97.13&97.13&97.13\\
gl\_ctg         & 98.48 & \bf \underline{98.93} &98.27 & 98.84&98.74&97.02&98.74&98.70\\
uk\_iu          & 97.03 & 97.12& \underline{97.25} & \bf 97.35&97.22&97.11&97.22&97.22$^\dagger$\\
eu\_bdt         & 96.48 & \underline{96.68} &96.66 &  \bf 96.71&96.63&96.33&96.63&96.62\\
da\_ddt         & 97.87 & \bf \underline{98.03} &97.74 & 97.95&97.87&97.57&97.91&97.87\\
sv\_talbanken   & 97.36 & \bf \underline{98.27} & 97.49 & 98.16 & 98.41 & 97.64 & 97.84 & 97.95\\
el\_gdt         & 96.66 & \bf \underline{97.38} &97.02 & 96.96&97.49&97.38&97.56&97.47\\
tr\_imst        & 97.03 & \bf \underline{97.39} &97.01 & 97.24&97.17&96.89&97.17&97.14\\
hy\_armtdp      & 95.55 & \bf \underline{96.01} &95.74 & 95.66&95.86&95.68&95.86&95.86$^\dagger$\\
be\_hse         & 81.91 &82.63& \bf \underline{83.33} & 82.92& 83.51&82.13&83.51& 83.51$^\dagger$\\
\midrule
Average&97.03& \underline{97.25} &97.17 & \bf 97.31&97.24&96.96&97.22&97.21\\
\bottomrule
\end{tabular}
\caption{Comparison of the enhanced models with the augmentation method: \mn{Def} is the \mn{Default} model, \mn{Apt} is the \mn{Apertium}-enhanced model, \mn{Def}+8K and \mn{Apt}+8K are the same \mn{Default} and \mn{Apertium}-enhanced models with augmented data. For the models marked with $\dagger$, the UD Lexicon is absent and is replaced with Apertium candidates instead.}
\label{tab:augmentation}
\vspace{-4mm}
\end{table*}

\begin{table*}[t]
\centering
\begin{tabular}{l | c c c| c c c}
\toprule
\bf Treebank&\bf \mn{Stanza} &\bf \mn{Stanza}+UD &\bf \mn{Stanza}+8K& \bf \mn{Apt} & \bf \mn{Lex}+UD & \bf \mn{Lex}+8K\\
\midrule
\midrule
 cs\_pdt & 98.58 & \bf \underline{98.76} & 98.60 & 98.49 & 98.70 & 98.66 \\ 
 ru\_syntagrus & 97.91 & 96.76$^\dagger$ & \underline{97.92} & \bf 97.98 & 97.36$^\dagger$ & 97.97 \\
 es\_ancora & 99.21 & \underline{99.25} & 99.15 & \bf 99.33 & 99.27 & 99.28 \\
 ca\_ancora & 98.49 & 98.29 & \underline{98.51} & \bf 98.85 & 98.83 & 98.83 \\
 fr\_gsd & 98.15 & 97.69 & \bf \underline{98.24} & 97.98 & 97.09 & 96.75\\
 hi\_hdtb & 96.66 & \underline{96.75} & 96.66 & \bf 98.84 & 98.76 & 98.71\\
 de\_gsd & \bf 96.78 & 97.53 & \underline{96.86} & 96.83 & 97.01 & 96.94 \\
 it\_isdt & 98.32 & \bf \underline{98.60} & 98.46 & 98.36 & 98.38 & 98.39 \\
 en\_ewt & 98.18 & \bf \underline{98.21} & 98.17 & \bf 98.21 & \bf 98.21 & 98.19 \\
 ro\_rrt & 98.16 & \bf \underline{98.44} & 98.27 & \bf 98.44 & 98.38 & 98.28 \\
 pt\_bosque & 98.12 & \bf \underline{98.32} & 98.12 & 98.30 & 97.98 & 98.20 \\
 nl\_alpino & 96.99 & \bf \underline{97.22} & 96.97 & 96.89 & 96.63 & 96.61 \\
 bg\_btb & \underline{97.36} & 96.26 & 96.62 & 98.02 & 98.12 & \bf 98.20 \\
 ur\_udtb & 95.62 & \underline{95.66} & 95.64 & 97.13 & 97.28 & \bf 97.29 \\
 gl\_ctg & 98.59 & \underline{98.64} & 98.60 & \bf 98.74 & 98.48 & 98.48\\
 eu\_bdt & \underline{96.52} & 96.51 & 96.41 & 96.63 & \bf 96.66 & 96.62\\
 da\_ddt & 97.36 & \bf \underline{97.89} & 97.55 & 97.87 & 97.82 & 97.70\\
 sv\_talbanken & 97.53 & \bf \underline{98.45} & 97.63 & 98.41 & 97.78 & 97.59 \\
 el\_gdt & 96.66 & 96.49 & \underline{96.89} & 97.49 & 97.52 & \bf 97.54 \\
 tr\_imst & 96.73 & \underline{96.90} & 96.83 & 97.17 & 97.17 & \bf 97.23 \\
\midrule
 Average & 97.60 & \underline{97.63} & 97.61 & \bf 98.00 & 97.87 & 97.87\\[1ex] 
\bottomrule
\end{tabular}
\caption{Evaluation of the effect of the \mn{Stanza}-based lexicon extension method;  comparison with the \mn{Apertium}-enhanced (\mn{Apt}) and the \mn{Lexicon}-enhanced systems (\mn{Lex+UD} and \mn{Lex+8K}).}
\label{tab:lexicons}
\vspace{-4mm}
\end{table*}

\subsection{Data Augmentation}
\label{sec:augmented}

We implemented the transducer augmentation method described by \citet{kanerva2020universal}. This method's basic idea relies on applying existing morphological analyzers (in this case, Apertium) to unannotated data to generate additional training instances. 
To obtain the augmentation data, we recreated the experiments of \citet{kanerva2020universal} with 8K additional data. First, we collected a word frequency list for each language based on automatically annotated CoNLL2017 corpora \cite{ginter2017conll}. For the languages not present in this dataset (Belarusian and Armenian), we used the wikidump to extract the word frequency list. Next, all words in the list were analyzed with the Apertium morphological analyzer. Then, we used the scripts\footnote{\url{https://github.com/jmnybl/universal-lemmatizer}} from the original experiments of \citet{kanerva2020universal} to convert the Apertium analyses to the UD format and filter out ambiguous cases. \textcolor{red}{Finally, the 8K most frequent words not already present in the training set together with their analyses were chosen and appended to the UD training set.}

Although both the enhanced and augmented systems utilize Apertium as the external source, additional data usage differs. The augmented system uses Apertium to create extra labeled training data, \textcolor{red}{while our enhanced model uses Apertium to generate additional lemma candidates to the words of the same initial training set.}
On the other hand, during test time, the augmented model must fully rely on the regularities learned during training, while our enhanced model can additionally look at the lemmas for words that were never seen during training. 

The comparison of our \mn{Apertium}-enhanced model and the augmented model is shown in the first two blocks of Table~\ref{tab:augmentation}. 
The first two columns reintroduce the \mn{Default} and \mn{Apertium}\textcolor{red}{-enhanced} models' results from the Table \ref{tab:main_results}, the third and the fourth columns show the same two models trained on the augmented training sets.
Overall, the average results for both \mn{Apertium}-enhanced and the augmented \mn{Default} model (the column \textsc{Def}+8K) are very similar, with the average of the \mn{Apertium}-enhanced model being slightly higher (97.25 vs. 97.17). The \mn{Apertium}-enhanced model is better in 15 languages out of 23 (underlined in the table), while the augmented model surpasses the enhanced model on 8 models. 
The \textsc{Apt}+8K column shows the results of a model combining both augmentation and enhanced methods\textcolor{red}{---the training data is first augmented with the additional 8K words and then additionally enhanced with the Apertium candidates via the second encoder.} The combined approach scores are the best for 8 languages out of 23, resulting in an average improvement over the \textcolor{red}{augmented \mn{Default} model of 0.14\% and over the \mn{Apertium}-enhanced model of 0.06\% in absolute.}  
These results show that both augmentation and enhancement methods can contribute in complementary ways.

\subsection{Lexicon Extension}

Another simple baseline method for using external data is to use a lexicon or an external system first and only resort to neural generation when the surface form (SF) is not present in the lexicon. This is essentially how the Stanza lemmatizer works. Stanza constructs a lexicon based on the training set. During inference, the prediction goes through a cascade of three steps: 1) if the SF is present in the lexicon, then the lemma is immediately retrieved from the lexicon. 2) If the SF is novel and is missing from the lexicon, an edit operation is generated that decides whether the SF itself or its lowered form is the lemma, or whether neither is true. 3) Only in the last case the lemma is generated by the sequential decoder.
For testing out the lexicon extension system, we used the pretrained \mn{Stanza} models but extended the lexicon stored in the Stanza system with additional items. Note that Stanza lexicons can only store one lemma per SF-POS combination. Thus, if any of the \textcolor{red}{external} lexicons contain ambiguous lemmas, the firstly encountered lemma is chosen for each word.

We extended the \mn{Stanza} lexicons with both the Apertium 8K datasets used for training the augmented models in section~\ref{sec:augmented} and the UD lexicons~\cite{sagot2018multilingual}. The results of these evaluations are shown in Table~\ref{tab:lexicons}. 
The set of languages in this table is slightly different than in Table~\ref{tab:main_results}, only including those languages for which the UD lexicons are existent.
The left block shows the results with various \mn{Stanza} models.
The first column shows the baseline \mn{Stanza} results (taken from Table~\ref{tab:main_results}), the second and the third columns present the \mn{Stanza} model with its lexicon extended with the UD lexicons and the 8K words, respectively. 
The original UD lexicon for Russian contained many erroneous lemmas due to poor post-processing, which skewed the average accuracy. Thus, we did additional post-processing to put it in line with other languages.

The average scores of the \mn{Stanza} systems extended with both UD and 8K lexicons remain roughly the same. However, when extending the \mn{Stanza} with UD lexicons, most languages improve at least slightly, as shown with the underlined scores in the column \mn{Stanza+UD}. 
Overall, on average, the simple lexicon extension method falls considerably behind our \mn{Apertium}-enhanced model (97.63 vs. 98.00), the scores of which are again replicated in the first column of the right-most block.

However, the \mn{Apertium}-enhanced model is not directly comparable to the \mn{Stanza} models with extended lexicons because 1) the training data differs as the enhanced model has access to extra lemma candidates of the training set words during training and 2) the lexicons available during the test time are different. 
Thus, we also show in the last two columns of the right-hand block of Table~\ref{tab:lexicons} the results of two \mn{Lexicon}-enhanced models \textcolor{red}{(recall Section~\ref{sec:experiments} and Table~\ref{tab:main_results})}, similarly extended with the UD and 8K lexicons. The \mn{Lexicon}-enhanced model has access to the same data as the \mn{Stanza} model during both training (training set + the training set based lexicon) and testing.

While the \mn{Lexicon}-enhanced model alone does not perform better than the \mn{Default} baseline (see results in Table \ref{tab:main_results}), adopting additional UD or 8K lexicons during test time increases the results to the same level with the \mn{Apertium}-enhanced model.
This shows that our proposed approach does not need additional resources during training---the model can be trained to use external sources based on the lexicon created from the training set. \textcolor{red}{Then, the system's real benefits can be achieved when using extra resources later during test time.} Without those resources, the model still performs on the same level as the non-enhanced baseline.

We hypothesize that our \textcolor{red}{dual-encoder approach} performs better than the \mn{Stanza} with extended lexicon \textcolor{red}{partly because of} the differences in the usage of the external data. Since \mn{Stanza} uses the lexicon resources as a first step in the cascade, it is prone to potential errors and noise in the lexicons. \textcolor{red}{The dual-encoder} model is safer against noise in this respect because the lemma candidates are not simply chosen as the prediction if present but are rather fed through the system that can decide how much to take or ignore from the given candidates.
Also, because \mn{Stanza} lexicons have the restriction of only one lemma per word-POS pair, the system might solve some ambiguities erroneously. Our approach is also more flexible in this respect, as the second encoder can be given several candidates, and again, the system learns to decide itself from which candidate how much to take.
\textcolor{red}{On average, there are $0.71$ lemma candidates per input word, and $1.09$ lemma candidates per input word when excluding those words that do not have external lemma candidates.}

\subsection{Effect of the Second Encoder}

Next, we performed a set of evaluations to argue for the effect of the \textcolor{red}{second encoder in the} enhanced model. We suggest that the improvements presented in Table~\ref{tab:main_results} for the \mn{Apertium}-enhanced model \textcolor{red}{over the \mn{Default} baseline} are indeed due to the input provided via the second encoder. To demonstrate that, we evaluated the test set for each language again, on the same model that was trained with Apertium lemma candidates but leaving the second encoder empty for the test time. 
For that, we retrained the \mn{Apertium}-enhanced models with the encoder dropout of $0.8$. This means that during training, 80\% of the time, the lemma candidates provided for the second encoder are dropped, and the model trains only the main encoder. The reasoning for using the dropout is similar to one provided for the \mn{Lexicon}-enhanced model in Section \ref{sec:method}---if the lemma candidates are always provided during training, the model learns to rely equally on both encoders. Due to that, if the second encoder remains empty during testing, the performance degrades considerably. If, on the other hand, the dropout is used, then the model learns to make predictions both when the candidates in the second encoder are present and also when they are absent. 
The results of these experiments are shown in the right-most block of Table~\ref{tab:augmentation}.

We first show in Table~\ref{tab:augmentation} that the results of the \mn{Apertium}-enhanced models trained with dropout are equivalent to the results obtained without dropout as evidenced by the column \textsc{Apt}$_{0.8}$.
Next, 
when the second encoder is empty (column \textsc{Apt}+E), the test results are similar to the ones obtained with the \mn{Default} model, providing evidence that the improvements are indeed due to the extra info supplied via the second encoder during test time.
Additionally, we emulated the scenario when extra lexicon information becomes available after training the model. In this case, it is straightforward to integrate this information into the system without having to retrain the model. The last two columns in Table~\ref{tab:augmentation} show the following scenarios on this respect: 1) Unimorph lexicons in addition to Apertium (7th column \textsc{Apt}+\textsc{Uni}) and 2) UD lexicons (the last column \textsc{Apt}+UD) in addition to Apertium.
The results in Table~\ref{tab:augmentation} show that, on average, extending the Apertium system with these particular lexicons do not add any benefit.
The reasons for that can be two-fold: 1) The UD lexicons are for most languages constructed based on the Apertium system and thus might not add any extra information; 2) The coverage of Unimorph lexicons in terms of lemmas is typically smaller than of Apertium systems. 

\textcolor{red}{Table~\ref{tab:examples} shows some examples when the \mn{Default} model predicted incorrect lemma while the \mn{Apertium}-enhanced model predicted the correct one.
In some cases, Apertium provided the only and correct candidate for the \mn{Apertium}-enhanced model, which was picked as a final prediction. In other cases, several candidates are provided to the second encoder, and the enhanced model chooses the correct one in most of the cases. This indicates that the second encoder effectively learns how to use the candidates to better control the lemma generation.}

\subsection{Effect of Morphological Features}

\begin{figure}[t]
\includegraphics[width=0.48\textwidth]{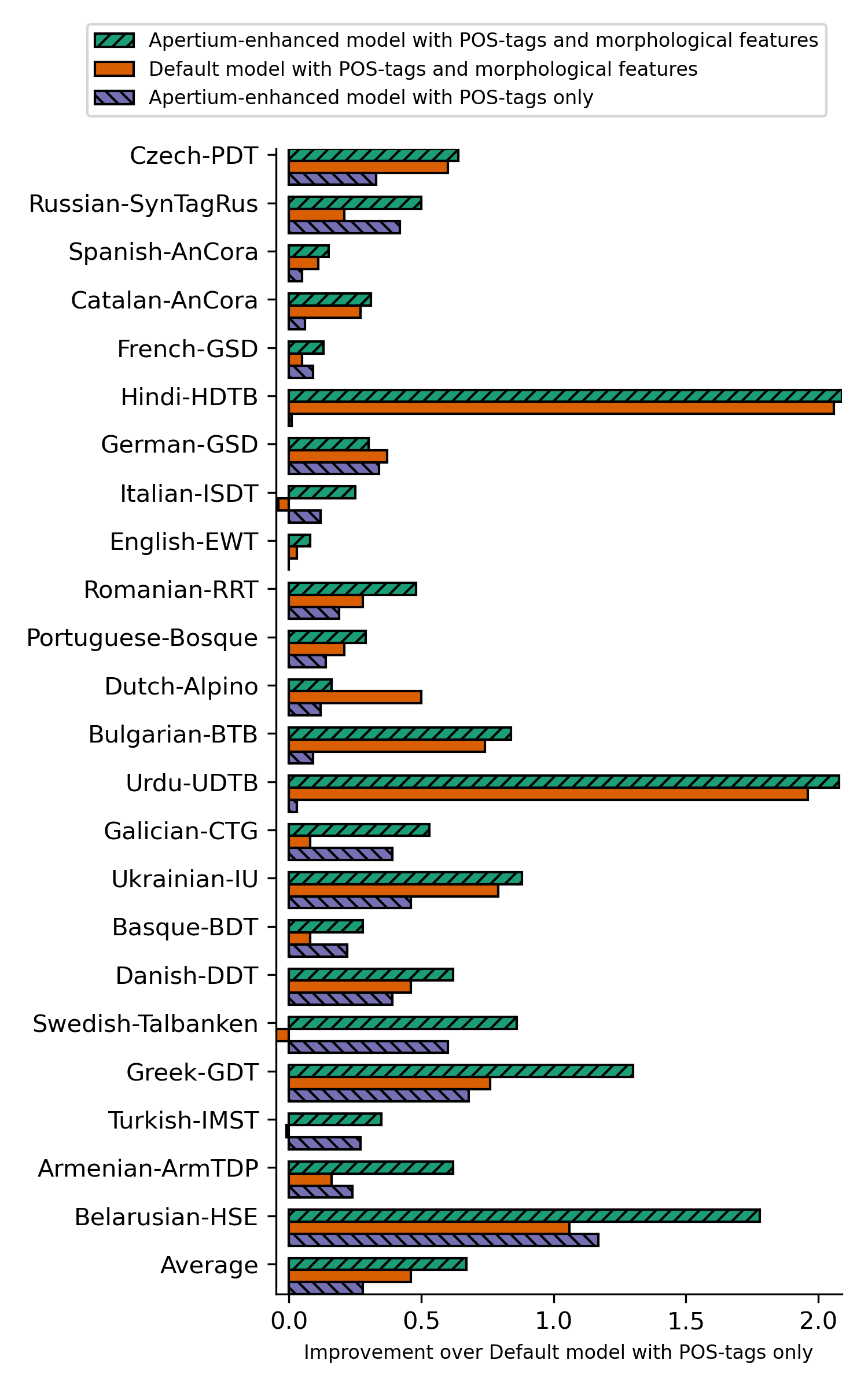}
\caption{Independent and cumulative effects of the second encoder and the morphological features on the model's performance. The origin of the x-axis is the performance of the \mn{Default} model with POS-tags only.}
\label{fig:pos_feats_diff}
\vspace{-4mm}
\end{figure}

{\color{red} 
All dual-encoder models were trained with both POS and morphological features in the input, while the \mn{Stanza} baseline only uses POS-tag information. Thus, the effect of the morphological features is a potential confounding factor when comparing the performance of the enhanced models to the \mn{Stanza} baseline.
To evaluate the effect of the morphological features, we trained the \mn{Default} and \mn{Apertium}-enhanced models with only providing POS-tag information to the input.

Figure~\ref{fig:pos_feats_diff} shows the improvement in accuracy over the \mn{Default} model trained with POS-tags only of 1) the \mn{Default} model trained with both POS-tags and morphological features, 2) the \mn{Apertium}-enhanced model trained with only POS-tags, and 3) the \mn{Apertium}-enhanced model trained with both POS-tags and morphological features. It can be seen that for some of the languages, the most improvement comes from adding morphological features to the input
, while for other languages adding the second encoder gives the main boost
.
However, for most languages,
combining the second encoder and morphological features provides the largest effect, which seems to be more complex than a linear combination of the two. We suppose that, in this scenario, the attention mechanism works differently---it allegedly takes the morphological features into account when picking the correct lemma from the multiple candidates.}

\begin{table}[t]
\small
\centering
\setlength\tabcolsep{3 pt}
\begin{tabular*}{\columnwidth}{ l l l l }
\toprule
\bf Input &\bf \mn{Def} &\bf \mn{Apt} & \bf Candidate(s) \\
\midrule
\midrule
  \foreignlanguage{russian}{папері} & *\foreignlanguage{russian}{папер} & \foreignlanguage{russian}{\textbf{папір}} & \foreignlanguage{russian}{папір} \\
  \emph{$\langle$paperi$\rangle$} & \emph{$\langle$paper$\rangle$} & \emph{$\langle$papir$\rangle$} & \emph{$\langle$papir$\rangle$} \\
  \foreignlanguage{russian}{чотирьох} & *\foreignlanguage{russian}{четвери} & \foreignlanguage{russian}{\textbf{чотири}} & \foreignlanguage{russian}{четверо, чотири} \\
  \emph{$\langle$\v{c}otyr'oh$\rangle$} & \emph{$\langle$\v{c}etvery$\rangle$} & \emph{$\langle$\v{c}otyry$\rangle$} & \emph{$\langle$\v{c}etvero, \v{c}otyry$\rangle$} \\
\midrule
  Antworten & *Antworte & \bf Antworten & antworten, antwort \\
  besten & bester & \bf gut & gut \\
\midrule
  \foreignlanguage{russian}{раскладзе} & *\foreignlanguage{russian}{раскладз} & \foreignlanguage{russian}{\textbf{расклад}} &  \foreignlanguage{russian}{раскласці, расклад} \\
  \emph{$\langle$raskladze$\rangle$} & \emph{$\langle$raskladz$\rangle$} & \emph{$\langle$rasklad$\rangle$} & \emph{$\langle$rasklasci, rasklad$\rangle$} \\
  \foreignlanguage{russian}{стаіць} & \foreignlanguage{russian}{стаіць} & \foreignlanguage{russian}{\textbf{стаяць}} & \foreignlanguage{russian}{стаяць, стаіць} \\
  \emph{$\langle$staic'$\rangle$} & \emph{$\langle$staic'$\rangle$} & \emph{$\langle$stajac'$\rangle$} & \emph{$\langle$stajac', staic'$\rangle$} \\
\bottomrule
\end{tabular*}
\caption{Examples for Ukrainian, German, and Belarusian words corrected by the enhanced model. All predictions of the \mn{Default} (\mn{Def}) are incorrect, the ungrammatical ones are marked with *. The correct predictions of the \mn{Apertium}-enhanced (\mn{Apt}) models are in \textbf{bold}. The last column shows the external candidates.}
\label{tab:examples}
\vspace{-4mm}
\end{table}

\section{Conclusion}
We proposed a method for \textcolor{red}{enhancing} neural lemmatization by integrating external input into the model via a second encoder and showed that the system incorporating Apertium morphological analyzer significantly improved the performance over the baselines. 
Both \citet{bergmanis2019training} and \citet{kanerva2020universal} used external resources to augment the training data, and thus, the improvement of their system is dependent on the amount and quality of the extended data supplied during training. 
On the other hand, our method trains the system to use the external information provided during run-time, thus making it independent of the particular external data available during training. 

We experimentally showed that the \textcolor{red}{enhancing method} is both slightly better and complementary to the data augmentation method of \citet{kanerva2020universal}.
We also compared our system with a simple lexicon extension method implemented based on the Stanza system. When trained and tested in a comparable setting, \textcolor{red}{the proposed enhanced} system achieves considerably higher results. 

{\color{red} Although the model's computational complexity is increased by introducing the second encoder, it is counterbalanced by our model being more robust to noise and the ambiguities stemming from the external lexicons.
Moreover, the main bottleneck in computation originates not from the neural network's increased size but can rather stem from the external system.
For example, in our experiments, the main bottleneck in computation originated from executing the transducer-based Apertium morphological analyser. To overcome this bottleneck, one possible trade-off between the speed and accuracy is to precompile a candidate list large enough to cover the most frequent words for a given language.}
This is a problem that also simpler baseline methods adopting external resources have to address.

Finally, it is worth noting that the proposed method could be beneficial for less-resourced languages. However, establishing this claim would need more systematic experiments exploring specifically on this question, which we did not focus on in this paper. Still, because the significant improvements shown in this work are obtained on languages with larger datasets, the possible gains on smaller datasets can be larger.

\section*{Acknowledgments}
The first author was supported by the IT Academy Program (StudyITin.ee).

\bibliographystyle{acl_natbib}
\bibliography{anthology,eacl2021}

\end{document}